%% file: arxiv-v1.tex
\documentclass[11pt]{article}
\usepackage{amsmath, amsthm, amsfonts, enumerate, amssymb, setspace, cite}
\usepackage[margin=1in]{geometry}
\usepackage{graphicx, cite}
\usepackage[dvipsnames]{xcolor}
\title{\textsc{Optimizing Closed-Loop Performance with Data from Similar Systems:\\ A Bayesian Meta-Learning Approach}}
\date{}
\author{Ankush Chakrabarty\thanks{Phone: +1 (617) 758-6175. Email: \texttt{achakrabarty@ieee.org}.  Mitsubishi Electric Research Laboratories (MERL), Cambridge, MA, USA.}}

\theoremstyle{definition}
\newtheorem{examplex}{Example}
\newenvironment{example}
{\pushQED{\qed}\examplex}
{\popQED\endexamplex}

\DeclareMathOperator*{\argmax}{arg\,max}

\usepackage{algorithm}
\usepackage{algpseudocode}
\newcommand\Algphase[1]{%
\vspace*{-.5\baselineskip}\Statex\hspace*{\dimexpr-\algorithmicindent-2pt\relax}\rule{\columnwidth}{0.1pt}%
\Statex\hspace*{-\algorithmicindent}\textbf{#1}%
\vspace*{-.5\baselineskip}\Statex\hspace*{\dimexpr-\algorithmicindent-2pt\relax}\rule{\columnwidth}{0.1pt}%
}
\newcommand\ignore[1]{{}}

\begin{document}
\maketitle

\begin{abstract}
Bayesian optimization (BO) has demonstrated potential for optimizing control performance in data-limited settings, especially for systems with unknown dynamics or unmodeled performance objectives. The BO algorithm efficiently trades-off exploration and exploitation by leveraging uncertainty estimates using surrogate models. These surrogates are usually learned using data collected from the target dynamical system to be optimized. Intuitively, the convergence rate of BO is better for surrogate models that can accurately predict the target system performance. In classical BO, initial surrogate models are constructed using very limited data points, and therefore rarely yield accurate predictions of system performance. In this paper, we propose the use of meta-learning to generate an initial surrogate model based on data collected from performance optimization tasks performed on a variety of systems that are different to the target system. To this end, we employ deep kernel networks (DKNs) which are simple to train and which comprise encoded Gaussian process models that integrate seamlessly with classical BO. The effectiveness of our proposed DKN-BO approach for speeding up control system performance optimization is demonstrated using a well-studied  nonlinear system with unknown dynamics and an unmodeled performance function.
\end{abstract}

\section{Introduction}\label{sec:intro}
Performance-driven optimization of closed-loop control systems often involves optimizing  system-specific performance functions. Some performance functions lack a complete mathematical representation, or exhibit such complex models to represent accurately, that standard model-based techniques cannot be deployed without oversimplifying assumptions or introducing conservativeness in the design.
As argued in~\cite{neumann2019data,lu2021bayesian}, Bayesian optimization~(BO) possesses some clear advantages that have led to its adoption for closed-loop performance optimization~\cite{duivenvoorden2017constrained, konig2020safety,bansal2017goal, paulson2020data}. Concretely, BO is a gradient-free, global optimization method that works well in data-limited settings while concurrently learning a probabilistic approximation of the underlying performance function from data. This has led BO to be implemented in a range of engineering applications, including energy  systems~\cite{chakrabarty2021boesc,xu2021vabo,khosravi2019controller}, vehicles~\cite{pal2020multi},  robotics~\cite{berkenkamp2021bayesian}, and aerospace~\cite{lam2018advances}. 

In the performance-driven BO literature, a  standard assumption is that the target system to be optimized is available for closed-loop experimentation. Therefore, at design time, one can evaluate the performance of the target system by changing some parameters such as controller gains or setpoints, and observing  the system response. Often, at the start of the BO tuning procedure,  the amount of data available from the target system is very limited. Approximating a performance function using very limited data generally results in an inaccurate surrogate model, which needs to be improved in subsequent BO iterations by exploring the search space~\cite{snoekBO_2012}. 
Though data available from the target system under consideration may be scarce, in many industrial applications there is often data collected from previous performance optimization tasks on `similar systems'~\cite{9867413}. We posit that this previously collected data, combined with the limited target system data, can warm-start BO by estimating a good initial surrogate model, potentially improving the convergence of BO for the target task.

To this end, we propose the use of few-shot meta-learning (or just meta-learning for brevity), wherein a machine learning model is trained on a variety of related `source' optimization tasks so that it can make accurate predictions for a new target task, in spite of the paucity of target task data. Modern meta-learning algorithms typically require solving a bi-level optimization problem, where the outer level is dedicated to extracting task-independent features across a set of source tasks, and the inner level is devoted to adapting to a specific optimization task with few iterations and limited data. This bi-level training loop is the basis of model-agnostic meta-learning (MAML)~\cite{pmlr-v70-finn17a}. However, MAML is not always easy to train and often exhibits numerical instabilities~\cite{antoniou2018train}. In addition, MAML does not generate uncertainty estimates around the predictions since it is a deterministic algorithm. Consequently, we adopt a Bayesian meta-learning approach in this work, which builds on the methodology proposed in~\cite{patacchiola2020bayesian,yoon2018bayesian,finn2018probabilistic}. To the best of our knowledge, this paper is the first to propose a Bayesian meta-learning algorithm for closed-loop system optimization capable of leveraging data from similar systems.

Our main contribution in this work is to propose a meta-learned Bayesian optimization methodology for rapidly optimizing closed-loop performance of systems with dynamics and performance functions whose mathematical representations are unknown, but can be evaluated by experiment/simulation. Specifically, we employ Bayesian meta-learning to systematically learn using data collected from a variety of systems that are different to a target system to be optimized, and leverage this disparate dataset to generate a good surrogate model for the target system performance function, despite having very few data samples from the target system. Our Bayesian meta-learning framework is implemented using deep kernel networks (DKNs) which have the advantage of requiring single-level optimization for training as the inner training loop is replaced by a Gaussian process (GP) base kernel. Additionally, our approach is simple to implement using standard deep learning toolkits such as \texttt{PyTorch}. The choice of GPs as the base learner lends itself to classical BO, where the default choice of probabilistic surrogate model is a GP. We demonstrate via simulation experiments that this performance-driven meta-learned BO can generate near-optimal solutions with fewer online experiments than classical BO for  systems with unmodeled dynamics, without requiring explicit parameter estimation.  While meta-learning has recently been explored in the context of adaptive control in~\cite{arcari2020meta,o2021meta,richards2021adaptive}, our work differs from these works in a few key aspects: (i) we propose a framework for meta-learning to optimize closed-loop performance rather than controller gain adaptation for tracking/regulation, and (ii) we adopt Bayesian deep learning for meta-learning and instead of adaptive controllers; and, (iii) we do not assume any modeling knowledge on the closed-loop dynamics, controller structure, or performance function.

The rest of the paper is organized as follows. The closed-loop performance optimization problem is formalized in Section~\ref{sec:psps} and the meta-learning setup is explained therein. In Section~\ref{sec:preliminaries}, we provide a brief overview of BO and deep kernel networks, and describe how they are used for meta-learned BO in Section~\ref{sec:metabo}. We demonstrate the effectiveness of the proposed approach using a numerical example in Section~\ref{sec:results}, and conclude in Section~\ref{sec:conc}.

\section{Motivation}\label{sec:psps}
We consider a class of stable closed-loop systems of the form
\begin{equation}\label{eq:cl_sys}
x_+ = f(x, r, \theta)
\end{equation}
where $x, x_+\in\mathbb R^{n_x}$ denote the current and updated state of the system, respectively, $r\in\mathcal R\subset \mathbb{R}^{n_r}$ is a control parameter to be tuned (\textit{e.g.}, a setpoint or controller gain), and $\theta\in\Theta\subset \mathbb R^{n_\theta}$ denotes unknown system parameters. The admissible sets of control parameters and system parameters: $\mathcal R$ and $\Theta$, respectively, are known. 
Additionally, we assume that for each $\theta\in\Theta$ and each $r\in\mathcal R$, the closed-loop system~\eqref{eq:cl_sys} is globally asymptotically stable to a parameter-dependent equilibrium state $x^\infty(r,\theta)$, and that the map $x^\infty(\cdot,\cdot)$ is continuous on $\mathcal R\times\Theta$.  

To determine how the closed-loop system performs, we define a continuous function $J:\mathcal R\times\Theta \to \mathbb R$. The performance output $J(r,\theta)$ is available for measurement. Note that for our purposes, it suffices that $f$ and $J$ are smooth and can be evaluated either by simulation or experiments---however, we do not assume access to a mathematical representation of these functions, nor can we compute their gradients analytically. In other words, no knowledge of $f$ or $J$ is used in the algorithm. However, $J$ is assumed to lie in the reproducing kernel Hilbert space (RKHS) of common (\textit{e.g.}, Gaussian or Mat\'{e}rn) kernels so that it can be approximated reasonably by kernel methods.

For a target system of the form~\eqref{eq:cl_sys} with unknown system parameter $\theta$, our \emph{target optimization task} is to compute the control parameter
\begin{equation}
\label{eq:objfn}
r^\star := \argmax_{r\in\mathcal R} J(r,\theta)
\end{equation} 
that optimizes the performance of the closed-loop system~\eqref{eq:cl_sys} directly from data. We will do this without estimating $\theta$, which would prove difficult since $f$ is unknown, and $x$ is not measured.

In practical applications that involve optimization of closed-loop performance, it is often the case that the optimization task has been performed before, most likely for similar, but not necessarily identical, systems. While solving these \textit{source optimization tasks}, one has likely generated optimization-relevant data that can prove valuable to the target optimization task, even if the target system is not exactly the same as one of the systems encountered before. In BO-based tuning, such source data (data obtained while solving source optimization tasks) is typically ignored, and the target system is optimized from scratch~\cite{neumann2019data}. This can lead to slow convergence of BO since the algorithm's initial estimate of the target performance function, constructed with very few target data points, is often inaccurate or highly uncertain. In this work, we propose meta-learning from similar systems' source data and integrating the information extracted from the source data to the target optimization task. In the sequel, we will show that meta-learning from disparate sources of data can enable optimization of closed-loop performance in a few-shot manner (i.e., with very limited target system data) and thereby lead to significant acceleration of the target optimization procedure.

\begin{example}
As a motivating example, consider a common task in sustainable building design: to minimize energy consumption of a newly constructed target building by tuning controllable variables in space heating/cooling systems~\cite{chakrabarty2021boesc}. At design time, it is likely that very little useful data has been collected from the target building, and using classical BO will require first exploring the search space until a good estimate of the energy function is learned. Conversely, we posit that the designer could have access to larger quantities of source data from other buildings that are similar in architectural style, location, occupancy, and HVAC equipment. Even if the energy functions learned from those source tasks are not identical to the energy function of the target building, a learner can extract information (\textit{e.g.}, curvature, regions in the parameter space likely to contain optima) about building energy functions, which can be used to rapidly optimize energy in the target building.
\end{example}

Concretely, we assume that a dataset $\mathcal D^S\triangleq \{\mathcal D^S_k\}_{k=1}^{N_S}$ is available, comprising data collected from $N_S\in\mathbb N$ source optimization tasks performed in the past. Here,
\begin{equation}\label{eq:kth_task_data}
	\mathcal D^S_k = \{r_{t,k}, J(r_{t,k},\theta_k)\}_{t=1}^{T_k}
\end{equation}
is a data sequence of $(r,J)$ pairs collected over $T_k\in\mathbb N$ optimization iterations from the $k$-th source task. Each $k$-th source data pair consists of the control parameter $r_{t,k}$, and the corresponding performance output $J(r_{t,k},\theta_k)$ evaluated on a source system with dynamics modeled by~\eqref{eq:cl_sys} with system parameter $\theta_k\in\Theta$, where each $\theta_k$ is unique. We reiterate that we do not assume access to the source systems or the system parameters $\theta_k$.
For the target task, our objective is to compute~\eqref{eq:objfn} by learning a variety of performance functions from the source dataset $\mathcal D^S$ and generating a good estimate of the performance function for the target optimization task, despite having access to a very limited target dataset $$\mathcal D^T=\{r_t, J(r_t,\theta)\}_{t=1}^{T},$$ where $T\ll \min_k T_k$. 

To this end, we propose the use of Bayesian meta-learning to learn a distribution of performance functions from the source data, and predict the target performance function by conditioning on the target data. With successful meta-learning, we expect that the estimated target task performance function will be significantly more accurate than an initial estimate that uses only the limited target dataset, and in turn, expect this to improve the convergence of BO. Our meta-learning framework comprises a deep kernel network (DKN). Deep kernel networks contain a latent encoder and a Gaussian process (GP) output layer. The latent encoder takes an input $r$ and transforms it to a latent variable of user-defined size, thereby lending itself to target systems with a large number of control parameters. Since the final layer is a GP regressor, it integrates seamlessly into classical BO procedures and well-studied acquisition functions can be used with the DKN without modification.

\section{Preliminaries}\label{sec:preliminaries}
In this section, we explain how classical Bayesian optimization is typically used to solve~\eqref{eq:objfn} for the closed-loop system~\eqref{eq:cl_sys} for a fixed $\theta$, using only target task data. We also describe deep kernel networks that will be used in the next section for meta-learning. 
\subsection{Bayesian Optimization (BO)}\label{subsec:vanilla_BO}
Classical BO comprises two main components: (1) a probabilistic surrogate model that maps $r\to J$ for $r\in\mathcal R$, and (2) an acquisition functions that balances exploration and exploitation to propose new candidate control parameters $r$ that are likely to optimize $J$, based on estimates of the surrogate model. Since classical BO starts with no initial data available for the system under consideration, one generally requires some random sampling on $\mathcal R$ to initialize a surrogate model. For each such control parameter $r_i$, $i\in \mathbb N$, the closed-loop system~\eqref{eq:cl_sys} is allowed to achieve a steady-state after some suitable wait time $t_f$, after which a measurement $J$ is taken. This steady-state performance output acts as a proxy for the objective function value for the corresponding $r_i$, that is, $J_i=J(t_0 + i t_f)$. This enables us to obtain an initial set of $(r_t, J_t)$ pairs from which a probabilistic surrogate model can be trained to estimate $J$ on $\mathcal R$.

In particular, classical BO methods utilize Gaussian process (GP) regression~\cite{williams2006gaussian} to construct the surrogate model. The Gaussian process framework works on the principle that the performance $J$ is a random variable and the joint distribution of all $J_i$ for the $n$ training data instances is a multivariate Gaussian distribution. This implies that the GP can be characterized entirely by a mean function $\mu:\mathcal R \to \mathbb R$ and a covariance function $K:\mathcal R\times\mathcal R\to \mathbb R$; that is,
$J \sim \mathcal N(\mu(r), K(r,r))$.
A common choice is to set $\mu\equiv 0$, and to represent the covariance function using kernels $K$ such as squared exponential kernels or Mat\'{e}rn kernels, which are parameterized by hyperparameters $\gamma$ such as length-scales and variances. At inference, for predicting the performance of a new instance $\hat r$, the assumption that the predictions are jointly Gaussian with respect to the training instances $\mathsf J = \begin{bmatrix} J_0 & \ldots & J_i & \ldots & J_n \end{bmatrix}^\top$ yields an estimate $\hat J$ with posterior mean and variance given by
\begin{equation}\label{eq:gp-mean-var}
\begin{split}
\mathbb{E}[\hat J] &=  \hat{\mathsf K}^\top \mathsf K^{-1} \mathsf J, \\
\mathrm{Var}[\hat J] &= \bar{\mathsf K} - \hat{\mathsf K}^\top \mathsf K^{-1}\hat{\mathsf K}
\end{split}
\end{equation}
where $\mathsf K = K(J, J|\gamma) + \sigma_w^2 I$ with $\sigma_w$ being an estimate of the variance of the noise corrupting the performance output measurement, $\hat{\mathsf K} = K(J, \hat J|\gamma)$, and $\bar{\mathsf K} = K(\hat J, \hat J|\gamma)$.
Clearly, the accuracy of the predictions depends on the choice of kernel and the kernel hyperparameters $\gamma$. In classical BO, $\gamma$ is computed by  maximizing a log-marginal likelihood function
\begin{equation}\label{eq:lml}
\mathcal L(\gamma) = -\frac{1}{2}\log|\bar{\mathsf K}| - \frac{1}{2} \mathsf J^\top \bar{\mathsf K} \mathsf J  + C_0,
\end{equation}
where $C_0$ is a constant.
Although this problem is non-convex in $\gamma$, one can employ stochastic gradient ascent to search for optimal hyperparameters.

The selection of a next best candidate $r$ in classical BO is performed via an acquisition function $\mathsf A:\mathcal R\to \mathbb R$. The acquisition function uses the
predictive distribution given by the GP to compute the utility of performing an evaluation of the objective at each $r_i$, given the data obtained thus far. The next $r_i$ at which the objective has to be evaluated is obtained by solving the maximization problem 
$r_{i+1} := \argmax_{\mathcal R} \mathsf A(r)$. 
Commonly used acquisition functions include expected improvement (EI) and upper confidence bound (UCB)~\cite{snoekBO_2012}.
With the new control parameter candidate $r_{i+1}$, we wait $t_f$ seconds to obtain $J_{i+1}$ and iteratively perform the  surrogate model retraining and the acquisition function evaluation steps with the training set augmented each BO iteration with the new pair $(r_{i+1}, J_{i+1})$. This process is repeated \emph{ad infinitum}.

\subsection{Deep Kernel Networks (DKNs)}\label{subsec:dkl}
While commonly used kernel functions have shown good performance in some applications, there are advantages to directly learning the kernel function from data. This is the idea proposed in deep kernel learning~\cite{wilson2016deep}, where a base kernel $K(\cdot, \cdot|\gamma)$ with hyperparameters $\gamma$ is used to transform the  control parameters to a latent space via a non-linear map $\mathcal F_\omega(r)$ induced by a deep neural network, parameterized by weights $\omega$.
\begin{figure*}[!ht]
    \centering
    \includegraphics[width=\columnwidth]{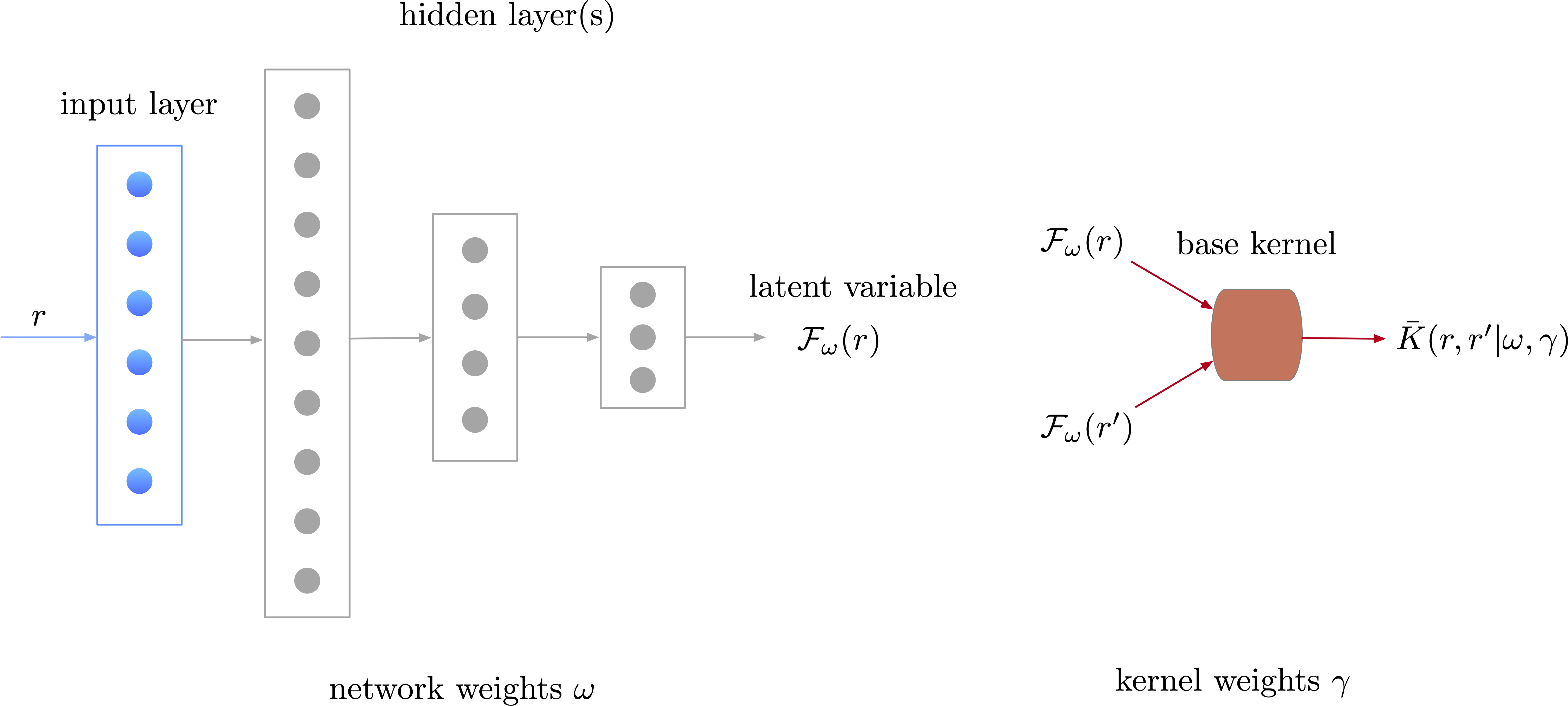}
    \caption{Architecture of the deep kernel network used for Bayesian meta-learning. The latent variable representation is learned by the neural network from multiple source tasks, and fixed during inference. The base kernel is used to estimate the mean and variance of a performance function.}
    \label{fig:DKN_architecture}
\end{figure*}

The deep kernel
\begin{equation}
\label{eq:dk}
\tilde K (r,r'|\gamma,\omega):= K\left(\mathcal F_\omega(r),\mathcal F_\omega(r')|\gamma\right) 
\end{equation}
is trained by maximizing a log marginal likelihood, similar to~\eqref{eq:lml}, to concurrently compute $\gamma$ and $\omega$. A noteworthy benefit of deep kernels is that they can be defined on a latent space whose dimension may be significantly lower than the dimensionality of the input, allowing the learning to be applied on problems where the dimensionality of the control parameters $n_r$ is large. A simple example of the DKN architecture is shown in Fig.~\ref{fig:DKN_architecture}.

\section{Meta-Learned Bayesian Optimization with Deep Kernel Networks (DKN-BO)}\label{sec:metabo}

Recall that $\mathcal D^S$ is a collection of data obtained from source tasks, where the data of the $k$-th source task is described in~\eqref{eq:kth_task_data}. Most recent Bayesian meta-learning paradigms~\cite{yoon2018bayesian,finn2018probabilistic} work under the assumption that all the source tasks (and the target task) share a set of task-independent hidden parameters $\omega$ and $\gamma$ (the reasons for using two symbols here will be explained shortly), and a set of task-specific parameters $\{\rho_k\}_{k=1}^{N_S}$. Given an input $r_*$ from an unseen task, the Bayesian treatment for inference requires estimating the task-independent parameters $\omega$ and $\gamma$ from the source task dataset, and then estimating two distributions: the posterior of the task-specific parameters $\pi(\rho_k|r_*, \mathcal D^S, \omega,\gamma)$ and the posterior for the output $\pi(J_*|r_*, \omega,\gamma)$. However, as argued in~\cite{patacchiola2020bayesian}, handling these two distributions requires sampling or amortized distributions, resulting in significantly complex network architectures, and bi-level learning has been shown to require higher-order derivative information, in spite of which the training might suffer from instabilities. By using deep kernel networks, one can marginalize out the task-specific parameters with kernel functions defined in the latent space, which leads to a simple architecture (see Fig.~\ref{fig:DKN_architecture}), allows for quantification of uncertainty during inference, all the while avoiding the numerical conditioning issues that are reported to occur~\cite{antoniou2018train} during training via bi-level optimization (\textit{e.g.}, in MAML).

\subsubsection*{Training}
The marginal likelihood of the Bayesian meta-learning module conditioned on $\omega$ and $\gamma$ is given by
\begin{equation}\label{eq:lml_bayes_meta}
\pi(J|r, \mathcal D^S, \omega, \gamma) = \prod_{k=1}^{N_S} \pi(J|r,\mathcal D^S_k,\omega,\gamma),
\end{equation}
where
\[
\pi(J|r,\mathcal D^S_k,\omega,\gamma) = \int \prod_{k=1}^{N_S} \pi(J|r,\mathcal D^S_k,\omega,\gamma,\rho_t)\,\mathrm{d}\rho_t
\]
represents the marginalization over sets of task-specific parameters. We approximate the Bayesian integral above  using a Gaussian process prior with kernel $\gamma$; this is a commonly used approach for tackling similar integrals that are implicitly solved via Bayesian quadrature methods, see for example,~\cite{courts2021variational,patacchiola2020bayesian}. 

Consequently, one can cast the meta-learning problem as learning the parameters $\omega$ and $\gamma$ that characterize a deep kernel network~\eqref{eq:dk}, which is tantamount to optimizing (maximizing) the log-marginal likelihood
\begin{equation}
\label{eq:DKN_loglikelihood}
\mathcal L_{\sf DKN}(\omega,\gamma) = \sum_{k=1}^{N_S} \left(-\frac{1}{2}\log|\tilde{\mathsf{K}}_k| - \frac{1}{2} \mathsf J_k^\top \tilde{\mathsf{K}}_k^{-1} \mathsf J_k \right) + C_1,
\end{equation}
where $\tilde{\mathsf{K}}_k := \tilde{\mathsf{K}}(r_k, r_k'|\gamma,\omega)$, $\mathsf J_k$ is a vector of all labels of the $k$-th source task, and $C_1$ is a constant scalar. 
In each training iteration, a source task is selected at random from the set of source tasks, and a random subset of batch data from this source task is selected. The log-marginal likelihood is computed using~\eqref{eq:lml_bayes_meta}, and well-known variants of stochastic gradient descent methods are used to minimize the negative of $\mathcal L_{\mathsf{DKN}}$. In other words, the learnable parameters are updated by back-propagation 
\begin{align*}
\omega &\leftarrow \omega + \beta_{\omega}\,\frac{\partial}{\partial \omega} \mathcal L_{\sf DKN}(\omega,\gamma),\\
\gamma &\leftarrow \gamma + \beta_{\gamma}\,\frac{\partial}{\partial \gamma} \mathcal L_{\sf DKN}(\omega,\gamma),
\end{align*}
where $\beta_\omega$ and $\beta_\gamma$ are the step sizes of the neural network component and the base kernel component of the deep kernel network, respectively. By optimizing the entire DKN in mini-batches, we can also limit the computational expenditure of the determinant and inverse operations during loss function evaluation, without losing guarantees of convergence of the training loss (up to a small neighborhood of a critical point of the loss function), as demonstrated in~\cite[Theorem 3.1]{chen2020stochastic}. We denote $\omega^\star$ and $\gamma^\star$ as the parameters of the DKN learned at the termination of the training loop.

\subsubsection*{Inference}
As in the few-shot optimization setting, we assume that very few data points $\mathcal D^T$ are initially available from a target task $T$. To predict the control performance $J$ on $\mathcal R$ for the target task, and subsequently compute a next candidate $r$ with Bayesian optimization, we rely primarily on the optimized task-independent parameters $\omega^\star$ and $\gamma^\star$, as the trained DKN is expected to have learned a suitable surrogate model of $r\mapsto J$ from the set of source tasks. By conditioning on the target data, we encourage the predictions of the DKN
\[
\pi(J_*|r_*, \mathcal D^T, \omega^\star, \gamma^\star)
\]
to be biased towards to the target task. Under the assumption that the target task is similar to the source tasks, we expect the predictions of the DKN conditioned on limited target task data to be more accurate and exhibit lower uncertainty than a regressor trained solely on the target data.

Analogous to the classical BO algorithm described in Section~\ref{subsec:vanilla_BO}, we  use the DKN surrogate to predict the control performance at samples in $\mathcal R$, and use these predictions to optimize an acquisition function. Optimizing the acquisition function yields the next best control input candidate, given the target task data collected so far. In subsequent DKN-BO iterations, even if $\omega^\star$ and $\gamma^\star$ are kept constant, since the kernel covariance matrix $\mathsf K$ is appended with new target task data (obtained by evaluating the performance $J$ of the target task with candidate control inputs $r$), the DKN predictions gradually tighten around the true performance function of the target task. A full pseudocode of DKN-BO is provided in Algorithm~\ref{algo:dkn-bo}.

The authors in~\cite{wistuba2021few} propose a method for introducing scale invariance during training of DKNs so that the target task labels do not need to be normalized to the unit line (which would be impossible, since the target task's maximum and minimum is unknown). In particular, the proposed method involves computing ${\mathsf J}_{\min}$ and $\mathsf J_{\max}$, which are the minimum and maximum performance output observed over all source tasks, respectively. Each time a batch is sampled during training, the labels are rescaled to $(J-{\mathsf J}_{\min})/({\mathsf J}_{\max}-{\mathsf J}_{\min})$, which, over a large number of training iterations allows the learner to learn trends from the source tasks that are scale independent, enabling inference for the unseen target task without  normalization.
\input{pc}

Empirically, we have noticed  improvements in DKN-BO performance when retraining the base kernel parameters $\gamma$ online (as in classical BO) while keeping the neural network weights, i.e. $\omega$, fixed. However, this approach is closer in spirit to transfer learning than meta-learning, and thus, we do not retrain the base kernel in the following subsection, to avoid online training expenditure. 
\section{Simulation Results}\label{sec:results}
We consider the following nonlinear system, previously studied in~\cite{guay2003adaptive,nesic2012framework}:
\begin{subequations}\label{eq:sim_sys}
\begin{align}
\dot x_1 &= x_2 - \theta_1 x_1,\\
\dot x_2 &= -\theta_2 x_1^2 + \kappa(x,r),\\
J &= 1 - \theta_1 x_1 - \theta_2 x_1^2,\label{eq:sim_sys_cost}
\end{align}
\end{subequations}
where
$\kappa(x,r) = -6x_1 + (\theta_1-5)(x_2-\theta_1x_1) + \theta_2 x_1^2 + r$,
is a control policy that renders the closed-loop system stable. For the target system, the true parameters are $\theta=[2,5]$ and the optimal control parameter is $r^\star=-1.2$ at which the system attains its maximum $J^\star=1.2$; this is not known to the designer. We choose $\mathcal R = [-10, 10]$ and $\Theta=[1, 6]^2$.

\begin{figure*}[!ht]
	\centering
	\includegraphics[width=\textwidth]{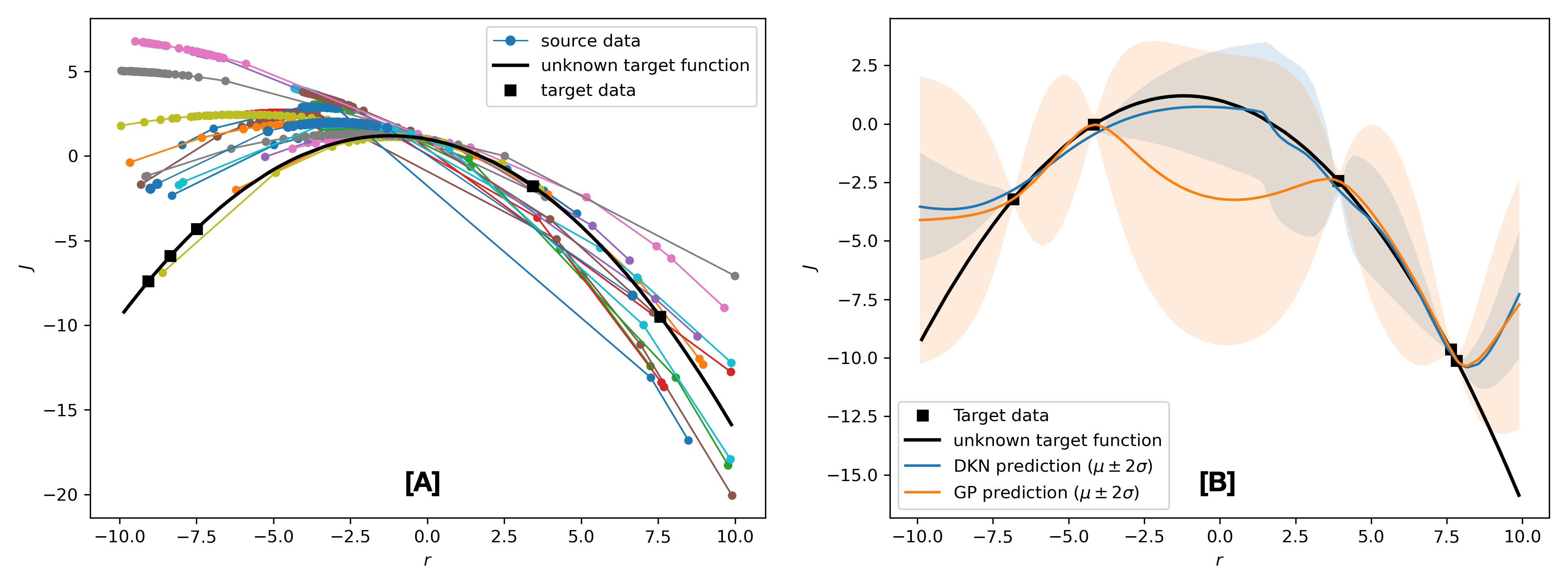}
	\caption{Data points (without noise) collected from source tasks shown with circles, limited data from target task shown with squares, and unknown target performance function shown with a continuous black line. Note that no target data point is at the optimal value.}
	\label{fig:data}
\end{figure*}
\begin{figure}[!ht]
	\centering
	\includegraphics[width=0.85\columnwidth]{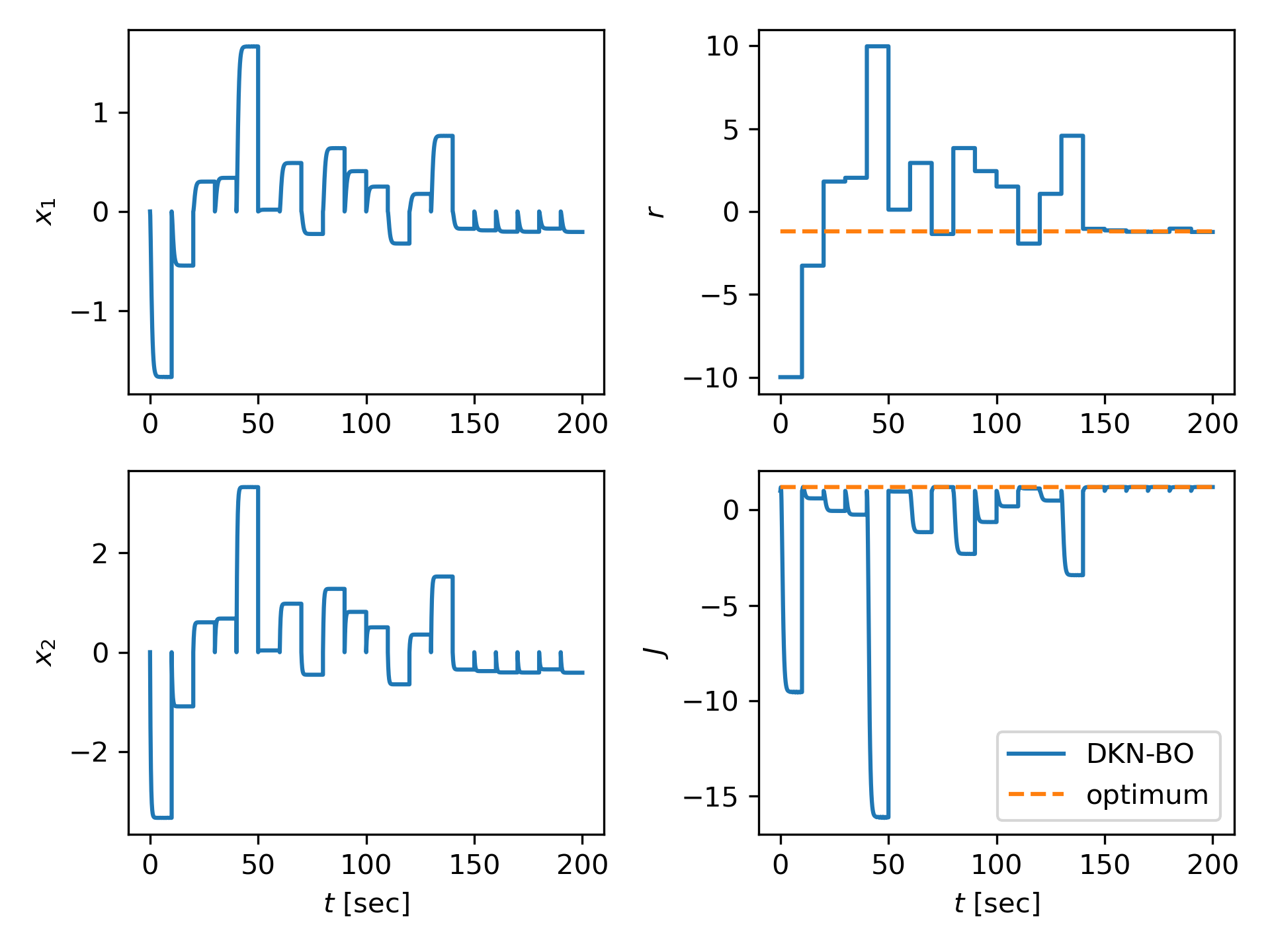}
	\caption{States $x$, controller parameter $r$, and performance output $J$ of the unknown target system during performance optimization by DKN-BO.}
	\label{fig:ctrl}
\end{figure}
\begin{figure}[!ht]
	\centering
	\includegraphics[width=0.8\textwidth]{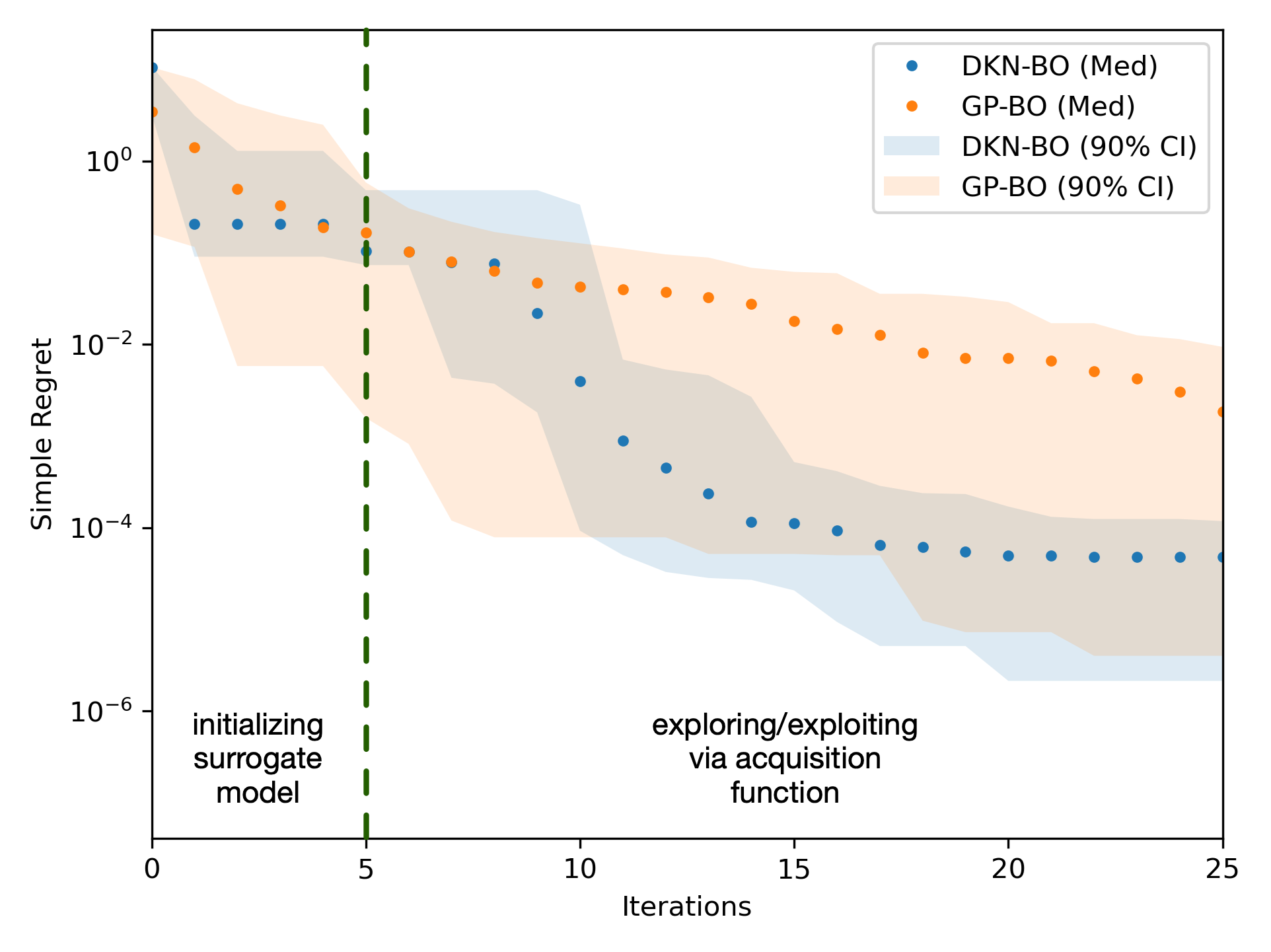}
	\caption{Regret decay (median and 90\% confidence interval) of target task's control performance cost for meta-learned DKN-BO (without base kernel retraining) and classical BO. The benefit of meta-learning is clearly seen through the fast convergence of DKN-BO.}
	\label{fig:regret-decay}
\end{figure}
In order to generate the source dataset, we extract $N_S=20$ unique system parameters from $\Theta$ using a 2D uniform random distribution. We evaluate the performance output $J$ for a given $r$ by forward simulating each system realized by sampling from $\Theta$ for $t_f=10~s$, which provides enough time for the transient response to fade. For each sampled system parameter, we solve the source optimization task of maximizing the performance function given in~\eqref{eq:sim_sys_cost} within $T_k\equiv 50$ iterations, which depends on $\theta$ and therefore has a variety of optimizers and optimal values. The source data collected is illustrated in Fig.~\ref{fig:data}[A] using colored circles joined by thin continuous lines. We observe that indeed the source performance functions are not identical to the target task performance function which is shown with the continuous thick black line. Our proposed approach is agnostic to the optimization strategy used for optimizing each source task; in fact, a mix of random search and BO are used for source task optimization and source data collection. The target task data is collected by evaluating the performance of the target system at $T=5$ randomly selected $r$ values.

Our DKN architecture involves a latent encoder with 100 neurons per hidden layer, with 4 hidden layers following the input layer. We select a latent dimension of 10, and the activation functions are rectified linear units (ReLUs). The latent output is passed to a GP with a scaled Matern-3/2 kernel as the covariance and a constant mean kernel. The entire pipeline is implemented using \texttt{PyTorch}~\cite{pytorchneurips} and \texttt{GPyTorch}~\cite{gardner2018gpytorch}. The DKN is trained for 10000 iterations with the learning rate scheduled at: $\beta_\omega=10^{-3}$ and $\beta_\gamma=10^{-2}$ for the first 2000 iterations, and $\beta_\omega=10^{-4}$ and $\beta_\gamma=10^{-3}$ thereafter. We train using the Adam algorithm and fix the batch-size at 8. The weights corresponding to the highest log marginal likelihood was saved as $\omega^\star$ and $\gamma^\star$.

Fig.~\ref{fig:data}[B] illustrates the effectiveness of meta-learning for predicting the target performance function. With the same limited target data, the trained DKN generates a more accurate mean estimate of the true performance function, as compared to a GP trained for 1000 iterations with a squared-exponential kernel: see the blue (DKN) and orange (GP) continuous lines. Furthermore, it is encouraging that the uncertainty bands for the DKN are less pronounced than the GP in the areas that are most relevant for optimization.

Fig.~\ref{fig:ctrl} demonstrates an instance of DKN-BO for optimizing control performance. We start the target system at the origin and simulate the system forward in time, collecting performance outputs every 10~s for $r$ value candidates generated by the DKN-BO procedure. The state variation $x_1$ and $x_2$ shows that we do not assume that we can reset the experiment each time a new $r$ candidate is selected, and the closed-loop performance is optimized in an online manner. The corresponding variation of the control parameter $r$. The bottom right subplot, which shows the performance output, illustrates clearly that  DKN-BO finds the optimal value within 100~s, which is comparable to the performance reported in~\cite{nesic2012framework}, even though we do not have access to any model information and do not estimate $\theta$.

We also repeated our experiments 100 times with varying target datasets, and report the statistics of the simple regret decay in Fig.~\ref{fig:regret-decay}. We observe that DKN-BO  performs well in a few-shot setting, and converges to a median regret that is two orders of magnitude below classical GP-BO within 10 iterations. Furthermore, the DKN-BO algorithm is robust to the target dataset as is evident from its tighter confidence intervals compared with GP-BO.

\section{Conclusions}\label{sec:conc}
This paper provides a highly generalizable and computationally simple framework for leveraging prior data for optimizing unseen closed-loop performance optimization tasks, even if the prior data is not from the task to be optimized. This is done in a few-shot manner using meta-learned Bayesian optimization and deep kernel networks as surrogate models. The latent encoding of the DKN enables a task-independent representation of the controlled inputs to the closed-loop system, and the base kernel allows conditioning for task-specific predictions.

\bibliographystyle{IEEEtran}
\bibliography{refs.bib}

\end{document}

%% file: pc.tex
\begin{algorithm}[htbp!]
	\caption{Meta-Learned Bayesian Optimization with Deep Kernel Networks (DKN-BO)}\label{algo:dkn-bo}
	\small
	\begin{algorithmic}[1]
	\Require $\mathcal R\leftarrow$ set of control inputs
    \Require $\mathcal D^S\leftarrow$ source task dataset
	\Require $\mathcal D^T\leftarrow$ target task dataset
	\Require $\beta_\omega, \beta_\gamma \leftarrow$  step sizes \Comment{default: $10^{-4}, 10^{-3}$}
	\Require latent vector dimension  \Comment{default: 8}
	\Require mini-batch size for training \Comment{default: 32}
	\Require number of training iterations \Comment{default: 10000}
	\Require $\omega,\gamma\leftarrow$ random initial DKN parameters 
	\Require $\mathcal{F}_\omega \leftarrow$ latent space encoder
	\Require $K(\cdot,\cdot|\gamma)\leftarrow$ base kernel
	\Algphase{Training the DKN}
	\For{each training iteration}
	\State $k\leftarrow$ choose random source task index
	\For{each mini-batch}
	\State sample random subsets of data from $\mathcal D^S_k$
	\State compute DKN loss using~\eqref{eq:DKN_loglikelihood}
	\State update $\omega$ and $\gamma$ using SGD/Adam
	\EndFor
	\EndFor
	\State $\omega^\star, \gamma^\star\leftarrow$ trained DKN weights
    \Algphase{Few-Shot BO}
    \State initialize surrogate model with trained DKN
    \For{each DKN-BO iteration}
    \State $\mu$, $\sigma\leftarrow$ predict performance function using DKN conditioned on $\mathcal D^T$
    \State evaluate acquisition function with predictions on $\mathcal R$
    \State $r_+\leftarrow$ next best BO candidate from acquisition function
    \State $J_+\leftarrow$ evaluate closed-loop performance with $r_+$
    \State Append target task dataset with $(r_+, J_+)$
    \EndFor
    \State \textbf{return} $r^\star\leftarrow$ control parameter for best performance observed
\end{algorithmic}
\end{algorithm}